\documentclass[12pt,a4paper]{article}
\usepackage[utf8]{inputenc}
\usepackage[english]{babel}
\usepackage{graphicx}
\usepackage{caption}
\usepackage{amsmath}
\usepackage[toc]{appendix}
\usepackage{amsfonts}
\usepackage{natbib}
\usepackage{xcolor} 
\usepackage{CJKutf8}
\usepackage{booktabs}
\usepackage{colortbl}
\usepackage{array}
\usepackage{xcolor}
\usepackage{graphicx} 
\usepackage{makecell}
\usepackage{caption}
\usepackage[colorlinks, citecolor=blue]{hyperref}

\title{Attention-aware semantic relevance predicting Chinese sentence reading }
\date{\vspace{-5ex}}
\author{
	\footnotesize{Kun Sun, University of Tübingen, Germany} \\
	\footnotesize{\texttt{kun.sun@uni-tuebingen.de}} 
}
\begin{document}

	\maketitle
	\begin{abstract}
In recent years, several influential computational models and metrics have been proposed to predict how humans comprehend and process sentence. One particularly promising approach is contextual semantic similarity. Inspired by the attention algorithm in Transformer and human memory mechanisms, this study proposes an ``attention-aware'' approach for computing contextual semantic relevance. This new approach takes into account the different contributions of contextual parts and the expectation effect, allowing it to incorporate contextual information fully. The attention-aware approach also facilitates the simulation of existing reading models and evaluate them. The resulting ``attention-aware'' metrics of semantic relevance can more accurately predict fixation durations in Chinese reading tasks recorded in an eye-tracking corpus than those calculated by existing approaches. The study's findings further provide strong support for the presence of semantic preview benefits in Chinese naturalistic reading. Furthermore, the attention-aware metrics of semantic relevance, being memory-based, possess high interpretability from both linguistic and cognitive standpoints, making them a valuable computational tool for modeling eye-movements in reading and further gaining insight into the process of language comprehension. 
Our approach underscores the potential of these metrics to advance our comprehension of how humans understand and process language, ultimately leading to a better understanding of language comprehension and processing.
		
	\end{abstract}
	\clearpage
	
	\section{Introduction}
	

	Researchers in cognitive sciences have proposed two influential theories regarding sentence comprehension and processing that have had a wide influence. The first theory is \textbf{expectation-based} and it proposes that the humans understand using expectations about the structure and content of sentence. According to this theory, language users actively predict what is coming next in a sentence and uses this information to guide the process of parsing and understanding the sentence. The expectation theory holds that expectations or predictions about upcoming content play a very important role in language comprehension and processing (e.g. \citealt{macdonald1994lexical}; \citealt{kutas2011thirty}; \citealt{huettig2015four}; \citealt{kuperberg2016we}. The computational metric, \textit{word surprisal}, has been shown to be a strong predictor of behavioural and neural measures for linguistic processing difficulty ( \citealp{demberg2008data}; \citealp{smith2013effect}; \citealp{hale2016information}; \citealp{shain2020fmri}). 
	
	The other theory is \textbf{memory-based} and it holds that humans rely on stored knowledge and information from their memory to guide the process of comprehending sentence (e.g., \citealp{baddeley2010working}). The theory explains that processing and comprehending language relies on stored semantic knowledge and that the relationships between linguistic units facilitate comprehension. This theory can account for a number of phenomena, such as the facilitation effect, namely that the processing of a word is faster and more accurate when it is preceded by semantically related words (e.g., \citealp{blank1978semantic}). Many studies have unveiled the presence of the semantic plausibility benefit effect in reading and language comprehension. However, many of these studies relied on human rating methods, where participants were asked to rate semantic relatedness or relevance to assess semantic plausibility (\citealp{hohenstein2010semantic}; \citealp{delong2014predictability}; \citealp{veldre2016semantic}). Nevertheless, the human rating approach is inherently imprecise and can be costly, making it impractical for large-scale text analysis. Because of this, semantic similarity (e.g., cosine similarity), as a computational metric, which concerns how closely related the meanings of two words or phrases are, can be used to make good predictions about language processing. Semantic similarity has been widely applied in NLP (natural language processing), cognitive psychology, and AI (see \citealp{harispe2015semantic}; \citealp{jabeen2020semantic}). A large body of empirical studies have demonstrated that semantic similarity is capable of predicting eye-movements on reading, and of predicting neural signals on language comprehension (\citealp{pereira2018toward}; \citealp{broderick2019semantic}; \citealp{hale2022neurocomputational}). The advantage lies in the ease of applying semantic similarity in processing large-scale texts, as compared to human rating. 

	At this point in time, a major task for researchers is proposing, developing and optimizing interpretable and highly-effective computational models/metrics in the computational cognition. 
	Contextual information is quite essential in computational models because it helps them to better understand and interpret the data they are working with. Without incorporating contextual information effectively, a model may make incorrect or biased predictions, especially when working with real-world data. For instance, using the \textit{attention} mechanism and ensemble (\citealp{vaswani2017attention}; \citealp{sagi2018ensemble}), deep learning models obtain far more contextual information and perform well. We believe that the computational models/metrics for cognitive sciences will become more robust after contextual information is effectively and fully acquired. Moreover, AI has grown exponentially, driven by successful deep learning models like RNN and Transformer. These models, inspired by natural cognition, have revolutionized computational methods. Concurrently, cognitive science has embraced computational approaches, viewing human cognition through this lens. The rapid advancement of AI and access to large datasets in cognitive and neural processing have fostered the development of computational cognitive science (\citealp{guest2021computational}; \citealp{taylor2021artificial}; \citealp{bartlett2023computational}).

	Word meanings can be represented using word vectors based on distributional semantics, such as those generated by word2vec programs or Transformers like BERT(\citealp{mikolov2013efficient}; \citealp{devlin2018bert}). Word vectors have been used in recent studies to investigate reading behavior using eye-tracking, EEG, and fMRI datasets. Most of these studies have relied on \texttt{correlation} statistical analysis to explore the relations between variables. However, correlation analysis is inadequate for understanding the effect of the word vectors on reading behavior  (\citealp{hollenstein2019cognival}; \citealp{hollenstein2021cmcl}). On the other hand, some studies have examined the predictive effect of semantic similarity on word or sentence processing (\citealp{mitchell2008predicting}; \citealp{mitchell2010composition}; \citealp{westbury2018conceptualizing}; \citealp{sun2022semantic}). In these studies, the relation between two word vectors is used to quantify the semantic relevance between the corresponding words. This approach allows for the use of more effective statistical methods for assessing effects of variables. While some studies have computed semantic similarity without taking context into account fully, others have advocated for the use of contextual semantic similarity to better capture how words are processed in different contexts. In order to compute more effective metrics, some researchers computed semantic similarity between a target word and its neighboring words in naturalistic discourse, that is, \textit{contextual semantic similarity}. This approach allows an individual word to obtain different values in different contexts, making it a more accurate predictor for reading behavior.
	
	Several approaches for computing contextual semantic similarity exist, such as the \textit{cosine} \citep{frank2017word} and \textit{Euclidean} methods \citep{broderick2019semantic}. However, such approaches may be improved in terms of comprehensiveness in the types of neighboring words considered and/or the linguistic interpretability of the semantic similarity values obtained. \citet{sun2023interpretable} proposed a \textit{dynamic} approach for computing contextual semantic similarity, which is cognitively and linguistically interpretable. Meanwhile, \citet{sun2023interpretable} modified the cosine \citep{frank2017word} and Euclidean methods \citep{broderick2019semantic}, and enabled them more interpretable and effective. However, the method of \citet{sun2023interpretable} ignored the different contribution of contextual words and the expectation effect. Based on the approach of \citet{sun2023interpretable} and inspired by the \textbf{attention} mechanism of Transformer \citep{vaswani2017attention}, the current study proposes a ``\textbf{attention-aware}'' method to compute more powerful and interpretable semantic relevance which can incorporate the contextual information more comprehensively. 
	
	Decades of research in human sentence processing have established that the time spent reading a word is an indicator of its processing difficulty. Therefore, reading time serves as a reliable measure of reading difficulty and can be recorded using various devices. Eye-tracking technology allows researchers to record the time spent on each word during reading, which is a reliable measure of processing difficulty. This has led to numerous studies that use eye-tracking data to test and improve computational models of language (\citealp{liversedge1998eye}; \citealp{rayner1998eye}; \citealp{rayner2006eye}; \citealp{schotter2014task}). The relationship between text and eye movements has prompted numerous studies that use eye-tracking data to enhance and test computational models of language (\citealp{barrett2015reading}; \citealp{demberg2008data}; \citealp{klerke2015reading}). We are interested in using eye-tracking databases as testing datasets to evaluate the effectiveness and validity of computational metrics proposed in the present study.
	
	 Further, the majority of computational metrics for language comprehension and processing have been primarily tested in English. Similarly, a significant portion of eye-tracking studies examining language comprehension and processing have been conducted using English as the primary language of study. However, new evidence indicates that the unique traits of the English language and the language habits of English speakers cause bias in the field, distorting research and leading to limited generalizations about all humans based solely on observations of English-speaking individuals \citep{blasi2022over}. 
	  Chinese is a logographic writing system, with each character representing a distinct morpheme. This is in contrast to alphabetic writing systems. The logographic nature of Chinese makes it a fascinating and challenging language to study. Researchers have conducted a multitude of experimental studies using eye-tracking and neural devices to investigate how native Chinese readers recognize and comprehend the language's complex characters (\citealp{yen2008eye}; \citealp{tan2000brain}). For instance, studies on eye-movements during Chinese reading have investigated various aspects of the reading process, including the role of morphological and semantic information, the effects of contextual information, and individual differences (\citealp{feldman1999semantic}; \citealp{taft1999positional}; \citealp{li2020integrated}). While there have been various studies investigating Chinese character or word recognition, there is a significant literature focused on eye movements during the reading of Chinese sentences (\citealp{liu2016effect}; \citealp{zang2018investigating}; \citealp{li2022universal}). Our interest lies in understanding how Chinese native readers process and comprehend naturalistic sentences or discourse during reading. In addition to these orthographic features, we are interested in how contextual semantic information influences word processing in Chinese reading. 
	 
	 Additionally, as is known, parafoveal-on-foveal effects in reading have been widely discussed. 
	 Specifically, the reader's visual system gathers information from neighboring words before the eyes land on them, and this information can affect how the reader processes the word they are currently looking at (\citealp{kennedy2005parafoveal}; \citealp{schotter2012parafoveal}). A debated topic was the existence of parafoveal-on-foveal effects in eye-movement control during reading. The E-Z Reader model proposed rare occurrences of such effects \citep{pollatsek2006tests}, while the SWIFT model suggested their prevalence due to parallel word processing \citep{engbert2005swift}. 
	 These differing perspectives on the mechanisms of reading have been subject to extensive debate and investigation in the field of cognitive sciences and psycholinguistics (\citealp{snell2018ob1}; \citealp{snell2019readers}). The debate has been informed by a wealth of empirical experiments. Such debates also exit in Chinese (\citealp{yan2012preview}; \citealp{liu2016effect}; \citealp{li2020integrated}; \citealp{li2023semantic}). Regardless these, we are also interested in whether the semantic relevance metrics we proposed can evaluate the relative advantages of each reading models. The current study is structured around two research questions, which are as follows:

~\\ 
\indent 1) Can the attention-aware semantic relevance metrics predict eye-movements in reading Chinese words?
~\\
~\\
\indent	2) How does the attention-aware metrics compare to other metrics computed by other approaches?
	
	\section{Related Work}
	
	\subsection{Word stroke count in Chinese reading}
	
	The present study will employ ``word stroke count'' as a predictive factor in statistical analysis, thereby requiring a brief introduction of the concept. In Chinese writing, a ``stroke'' refers to a single, uninterrupted movement of the writing instrument (such as a brush or pen) used to create a character. The number of strokes in a Chinese character can range from 1 to as many as 64, depending on the complexity of the character. In Chinese writing, the \textbf{stroke count of a word} is determined by adding together the stroke counts of the individual characters that make up the word. For example, the word for ``apple'' in Chinese is \begin{CJK*}{UTF8}{gbsn}苹果\end{CJK*} (píng guǒ), which is made up of two characters: \begin{CJK*}{UTF8}{gbsn}苹\end{CJK*} (píng) and \begin{CJK*}{UTF8}{gbsn}果\end{CJK*} (guǒ). \textbf{The first character} \begin{CJK*}{UTF8}{gbsn}苹\end{CJK*} has 8 strokes, and \begin{CJK*}{UTF8}{gbsn}果\end{CJK*}, the second character, has 8 strokes either, so the word \begin{CJK*}{UTF8}{gbsn}苹果\end{CJK*} has a total of 16 strokes. The other one-character word \begin{CJK*}{UTF8}{gbsn}吃 (chī) (``eat'')\end{CJK*} has 6 stokes.  Similarly, the stroke count of a Chinese sentence or phrase can be determined by adding together the stroke counts of all the characters in the sentence or phrase. Additionally, \textit{word length} usually refers to the alphabetic letter number for a word. After the concept of word length is introduced to Chinese, its basic meaning has been changed. In Chinese, word length used in the relevant studies actually refers to the number of character in a unit rather than alphabetic letter number. 


Existing work looks at and controls for effects of character stroke count during naturalistic reading of Chinese. 
These studies show that the stroke count of an individual character is an important factor during Chinese reading (\citealp{chen1996functional}; \citealp{wu2012erps}). For instance, \citet{li2014reading} found that characters with fewer strokes usually have shorter fixation durations when they are fixated. However, the study of \citet{li2014reading} only examined the stroke count for an individual character rather than the stroke count for a word. 
Although some studies examined how stroke count of an individual character influenced Chinese word/character recognition, quite rare studies have investigated what role of stroke count for word plays in Chinese naturalistic discourse/sentence reading. Among the top 10,000 Chinese words, the majority (78\%) are disyllabic words (two-character word), while 15\% of these words consist of only one character. Using the `stroke count of character' alone may be inadequate for providing comprehensive information to explain the processing of two-character words, especially in the context of naturalistic reading. Compared with ``word length'', ``word stroke count'' has some advantages. We therefore propose that the ``stroke count of a word'' could serve as an effective predictor in Chinese reading, complementing the role of ``word length''. The specific evidence and analysis is seen in the Appendix.

	\subsection{Semantic similarity based on word vectors in cognitive studies}
	
	Word vectors were used to represent the meanings and contexts of target words in previous research, where dimensions corresponded to words in the vocabulary or documents in a collection. Recent studies have increasingly adopted word embeddings obtained through neural language models (\citealp{mikolov2013efficient}; \citealp{bojanowski2017enriching}; \citealp{devlin2018bert}), which are more effective in capturing the semantic and grammatical information of words. 
	The relationship between word vectors and word procesing in reading comprehension has been well documented. Researchers have identified neural correlates between word vectors in both single-word comprehension and narrative reading, indicating the usefulness of word vectors in predicting reading behavior (\citealp{mitchell2008predicting};  \citealp{wehbe2014aligning}). With the advancement of NLP techniques and the availability of experimental databases on language comprehension, plentiful studies studies have explored the feasibility of using word embeddings to predict reading behavior (\citealp{hollenstein2019cognival}; \citealp{hollenstein2021cmcl}). 

 The discussed studies have typically used \texttt{correlation} statistical analysis to investigate the relation between word vectors and reading behavior. However, correlation as a basic statistical technique is limited in testing the effect of a variable on the response variable, compared to regression analysis.  In contrast, advanced regression analysis enables the examination of random effects on the response variable, a task that cannot be accomplished with the basic correlation technique. And it further allows for the comparison of the effects of various variables on the response variable.  To employ regression analysis in examining the effect of word vectors for reading behavior, some researchers have transformed word vectors into semantic similarity metrics (\citealp{mandera2017explaining}; \citealp{hollis2016principals}). However, fixed semantic similarity metrics for isolated words are insufficient in capturing the varying processing of the same words in different contexts during naturalistic discourse comprehension. Recent studies have used contextual semantic similarity metrics obtained from word embeddings to predict word processing in real-world naturalistic discourse processing tasks, acknowledging that the semantic context can influence word processing (\citealp{frank2017word}; \citealp{broderick2019semantic}). Contextual semantic similarity calculate for each instance of a target word in naturalistic discourse, taking into account its contextual or neighboring words. The following discusses the methods for how to compute it.

	 The \textit{cosine} approach has been widely used to calculate semantic similarity in studies of human word processing. In a study by \citet{frank2017word}, the contextual semantic similarity between a target word and its preceding content words within a sentence was computed using cosine similarity. To compute the metric, the concatenation of the vectors of the preceding content words was used as a new vector. However, this approach suffers from low interpretability, as it is unclear what the summation of the vectors of content words preceding a target word represents mathematically or linguistically. Furthermore, the exclusion of function words may not be justified, as function words can also affect the processing of the target word. Moreover, the \textit{Euclidean} approach of \citet{broderick2019semantic} can also be improved. First, the window size of the preceding words varies greatly. Second, the approach for averaging the preceding word embeddings is not given a mathematical and cognitive interpretation. This problem also exits in the cosine approach of \citet{frank2017word} where the preceding vectors are in summation. For more information, please refer to \citep{sun2023interpretable}. In order to obtain effective and interpretable metrics, \citet{sun2023interpretable} proposed a \textit{dynamic} method and modified the methods of computing aforementioned contextual semantic relevance. The \textit{dynamic} method, however, still has potential for improvement, which will be further discussed in the next section.

	\subsection{Attention mechanism and contextual information}
	
	Attention has been extensively studied in neuroscience and psychology, with research exploring its relationship to a variety of topics such as awareness, vigilance, saliency, executive control, and learning. However, few studies on the role of attention in human language processing have been done (\citealp{tomlin1999mapping}; \citealp{talmy1996windowing}; \citealp{myachykov2005attention}; \citealp{mishra2009interaction}). 
	We propose some methods to compute and evaluate the information of attention in language or language processing to some extent, and further making theoretical and experimental contributions. Attention could be a limited capacity processing system that can allocate resources in a flexible manner to modulate signal detection and response for controlled action. Regarding human language processing, attention plays a crucial role in selecting among competing options to activate the correct linguistic units (\citealp{kurland2011role}; \citealp{divjak2019frequency}).  Moreover, attention and memory could work together to enable us to learn, retain, and retrieve information. Attention helps us to encode, consolidate, and retrieve memories, and can be optimized through the use of strategies such as focused attention and minimizing distractions \citep{oberauer2019working}.
	
	In recent six years, attention has found increasing use in various domains of deep learning \citep{lindsay2020attention}, despite the fact that it does not have the closest ties to biology and psychology. For instance, attention is a mechanism used in the Transformer model to help it process input sequences. The Transformer is a neural network architecture that was introduced by \citet{vaswani2017attention}. The Transformer has been widely used in NLP tasks \citep{wolf2020transformers}. The attention mechanism used in the Transformers is a highly effective computational tool for computing contextual information in sequences, such as text.
	
	The attention mechanism allows the Transformer to focus on different parts of the input sequence at different times. This is done by calculating a set of weights, which indicate the importance of each input element for each output element effects (A simplified version of the attention mechanism can be found in the left panel of Fig.\ref{fig:attention}). The attention mechanism in the Transformer is highly effective in capturing long-range dependencies in input sequences (cf. \citealp{niu2021review}). Nevertheless, the attention weights are determined through a similarity function that compares each input element to each output element and are trained and then optimized in neural networks, resulting in their lack of interpretability. Consequently, In this study, directly applying the attention algorithm from the Transformer model is not feasible. However, this limitation offers an opportunity to explore innovative approaches for metric computation. By extracting contextual information, we can develop new methods that are informed by linguistic and cognitive insights. 

	Moreover, the \textit{dynamic} method for calculating contextual semantic similarity \citep{sun2023interpretable} is actually consistent with the fundamental idea of the attention mechanism in Transformer, but there is room for improvement. To enhance the performance of computational metrics, the current study proposes expanding the contextual window beyond just the preceding words, which serve as a type of memory, to include the following word that represent \textit{expectation}. By incorporating both memory-based and expectation-based strengths, we can develop new metrics that more accurately reflect the underlying meaning. Additionally, we suggest considering \textit{weights} based on the distances of surrounding words from the target word because the contributions of neighboring words can vary depending on their distance from the target word. Weights play a role of attention here in memory-based information. 
	Weights allow for obtaining the relevant information and ignoring irrelevant information. This ability to selectively attend to relevant information can help capturing the contextula information more effectively and accurately. 

	\subsection{Reading models}
	
	The debate on eye movements during reading has been a topic of significant interest in the field of cognitive psychology and psycholinguistics. Two prominent models that have been widely discussed and compared are the E-Z Reader and the SWIFT model (\citealp{pollatsek2006tests}; \citealp{engbert2005swift}). The E-Z Reader model posits a serial processing mechanism during reading, wherein words are sequentially recognized from left to right, akin to the sequential processing of letters in an alphabetic sequence. According to this model, fixations on words are triggered by specific visual input, and the reading process entails the sequential identification of individual words. In contrast, the SWIFT model presents an alternative perspective. This model contends that the reading process involves the integration of information from a broader visual span, not strictly confined to processing single words at a time. The SWIFT model proposes that readers may opt to skip certain words during fixations and utilize contextual cues to facilitate word recognition, thereby enabling a more flexible and efficient reading process. The debate between the two models centers on how exactly eye movements are controlled during reading and how words are recognized and processed in the reading task. Evidence from corpus studies supported the SWIFT model, while sentence studies showed limited support, aligning with the E-Z Reader model. Moreover, the E-Z Reader model posits serial processing during reading, while the SWIFT model contends that language processing occurs in parallel. The critical point between the two models is what role of \textbf{preview} plays in reading. While the E-Z Reader and SWIFT model represent contrasting viewpoints, research in this area is continually evolving, and some studies may even propose hybrid models that incorporate elements from both perspectives \citep{wen2019parallel}. 
	
	Parafoveal-on-foveal effects (i.e., preview benefits) in reading refer to the influence of information from the words located in the parafoveal region (the area immediately surrounding the fixated word) on the processing of the currently fixated word (the foveal word). Reading time on the target word is shorter when it is identical to the preview word, compared to when they differ, indicating that the preview word is processed with parafoveal vision. This preview effect has also been found in Chinese reading. For instance, preview benefits have been detected when the preview word and the target word share phonological information (\citealp{liu2002use}; \citealp{tsai2004use}; \citealp{yen2008eye}). These findings strongly suggest that words can be processed to a certain degree, or even fully, with parafoveal vision. Additionally, these studies provide evidence that at least some Chinese words are processed in parallel, supporting the notion of parallel word processing during reading.

	Moreover, in English reading, researchers have found the significance of semantic preview plausibility for parafoveal processing (\citealp{schotter2016semantic}; \citealp{veldre2016semantic}; \citealp{antunez2022semantic}). Various studies have revealed that semantically related or highly similar preview words lead to reduced fixation durations on the target word compared to semantically unrelated preview words, demonstrating the semantic plausibility benefit effect (\citealp{hohenstein2010semantic}; \citealp{schotter2013synonyms}). In the context of reading Chinese, some studies have explored how the semantic plausibility of the preview word influences the processing of the target word (\citealp{yang2012semantic}; \citealp{li2023semantic}). However, a limitation of these studies is that they relied on participant ratings to determine semantic relatedness to estimate semantic plausibility. This approach may not be feasible for large amounts of stimuli texts, and it lacks a quantifiable metric. With the advancements in NLP and AI techniques, we now have a range of methods to compute semantic relevance with greater precision for massive texts, eliminating the need for human ratings \citep{sun2023interpretable}. As previously discussed, considering the limitations of existing methods for computing semantic relevance, we propose the implementation of an ``attention-aware'' method to achieve more accurate and robust computations of semantic relevance. This method could provide a more robust and efficient approach to evaluate semantic plausibility in reading Chinese based on a massive dataset.
	
	Further, many studies have demonstrated that prediction plays a vital role in guiding eye movements during reading. For instance, \texttt{word surprisal}, an expectation metric, is widely used to show its significant effect on reading behavior (\citealp{hale2001probabilistic}; \citealp{levy2008expectation}).  Specifically, word surprisal, defined as the negative logarithm of word probability (i.e., $surprisal(w) = -\log_2 P(w|context)$, w = tagert word
			), has gained popularity in predicting English reading time (\citealp{hale2001probabilistic}; \citealp{levy2008expectation}; \citealp{demberg2008data}; \citealp{monsalve2012lexical}; \citealp{goodkind2018predictive}). Surprisal underscores the effect of prediction by considering only the left context, thereby supporting a serial processing model. Conversely, parafoveal processing captures information from the upcoming word, potentially conflicting with expected predictions. However, parafoveal processing provides only a preview of a small portion of the upcoming word, enabling the partial processing of its attributes, with its full identity remaining uncertain until fixation. During this process, readers may consistently predict forthcoming words based on contextual and linguistic cues. We will discuss the debate in the present study.

	
		Considering the limitations and insights provided by previous studies, this current study aims to achieve several objectives. First, we aim to introduce the ``attention-aware'' approach for calculating contextual semantic relevance for a target word. This approach will be both cognitively and linguistically interpretable, which is essential given the current focus on model interpretability in recent machine learning and computation research (\citealp{biecek2021explanatory}; \citealp{thampi2022interpretable}). Second, we will evaluate the effectiveness of these attention-aware metrics by using a large-scale dataset of eye-movements from Chinese reading with naturalistic discourse/sentence. Third, we assess whether the preview benefit can be observed in employing various metrics we proposed to predict eye-movements on Chinese reading. Forth, we use advanced regression methods to compare the performance of our metrics with other metrics computed by the existing approaches.
	
	\section{Materials and Methods }
	
	\subsection{Materials}

 This study used data from the dataset of eye-movement measures on words in Chinese reading, which is a corpus of eye-tracking data that includes predictability norms \citep{zhang2022database}. There are other several reasons why this corpus was chosen. First, the dataset employs sentences as stimuli. Specifically, the stimuli consist of 7577 Chinese sentences that collectively contain over 8000 different Chinese words. Second, the dataset recruited a large number of Chinese native speakers (n = 1718) to complete the reading tasks in 57 experiments. Third, the dataset includes data on different types of eye-movement measures in reading. The present study focuses on three specific eye-movement metrics as response variables, namely, \textit{first duration}, which refers to the duration of only the first fixation on the target word, \textit{gaze duration}, which refers to the sum of all fixations on a word prior to moving to another word, and \textit{total duration}, which refers to the summed duration of all fixations on the target word. Fourth, the datset is the largest eye-tracking one on Chinese reading, providing a rich source of data for analysis. Finally, the choice of an eye-movement database over an EEG or fMRI database was motivated by the fact that the measures in eye movements on reading and is more amenable to the statistical models used to assess and compare different metrics in the present study.

	Considering the objective of the current study and informed by the successful use of word2vec word embeddings in prediction tasks, we employed a set of one-million Tencent pretrained word2vec-style Chinese word embeddings to train the various approaches to semantic relevance. These word embeddings, available at \url{https://ai.tencent.com/ailab/nlp/en/download.html} (200 dimensions, large)\citep{song2018directional}, were pre-trained word2vec-style database including 12 million Chinese words and phrases. 
	According to \citet{song2018directional},  the superiority of Tencent corpora mainly lies in coverage, freshness, and accuracy, compared to existing embedding corpora. 

	\subsection{Methods}
	This study proposed a new ``attention-aware'' approach to computing contextual semantic relevance and compares it to two existing approaches, namely, the \textit{cosine} method modified from \citet{frank2017word} and the \textit{dynamic} method from \citet{sun2023interpretable}, all trained on the same pre-trained database of word embeddings. The \textit{cosine} method proposed by \citet{frank2017word} only considered content words. To improve its effectiveness, \citet{sun2023interpretable} introduced a modified \textit{cosine} approach, in which \citet{sun2023interpretable} considered both content and function contextual words. Given a target word, the vectors of the three preceding words are concatenated regardless of whether they are content or function words. \citet{sun2023interpretable} then used the same \textit{cosine} approach to process the two rows of vectors representing the target word and the sum of the vectors of the preceding three words. This value is taken as the \textit{cosine semantic similarity }value of the target word. As detailed in \citet{sun2023interpretable}, the window size of context may vary greatly in the Euclidean approach proposed by \citet{broderick2019semantic}. However, the Euclidean method is greatly influenced by high-dimension vectors, and its performance is unstable. Although the \textit{cosine} and \textit{Euclidean} approaches have been partly optimized, their inherent weaknesses still remain. Due to these issues, we merely chose to compare the metric computed by the modified \textit{cosine} method with the new metrics proposed in the present study. The following outlines how to compute these new attention-aware metrics.
	
	To compute the attention-aware metrics, we utilized the text stimuli from the Chinese reading eye-tracking corpus and extract vectors from pre-trained word embeddings that represent the target word and its surrounding words. In contrast to the \textit{dynamic} approach by \citet{sun2023interpretable}, we incorporate both the preceding and following words to better capture the information from the context. Additionally, we introduced weights based on the distance between the target word and its surrounding words to account for varying contributions from the contextual words. 
	This new approach takes into account both memory-based and expectation-based strengths, and weighs the different contributions from the contextual words as well.
	
\begin{figure}[ht]
	\centering
	\begin{minipage}[b]{0.29\linewidth}
		\includegraphics[width=\linewidth]{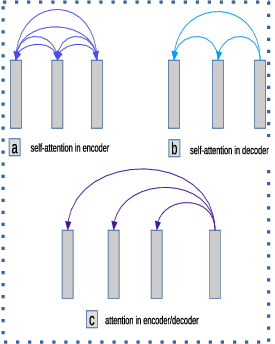}
		\caption*{\footnotesize{\textbf{A}. Attention types in Transformers}}
		\label{fig:imageA}
	\end{minipage}
	\hfill
	\begin{minipage}[b]{0.70\linewidth}
		\includegraphics[width=\linewidth]{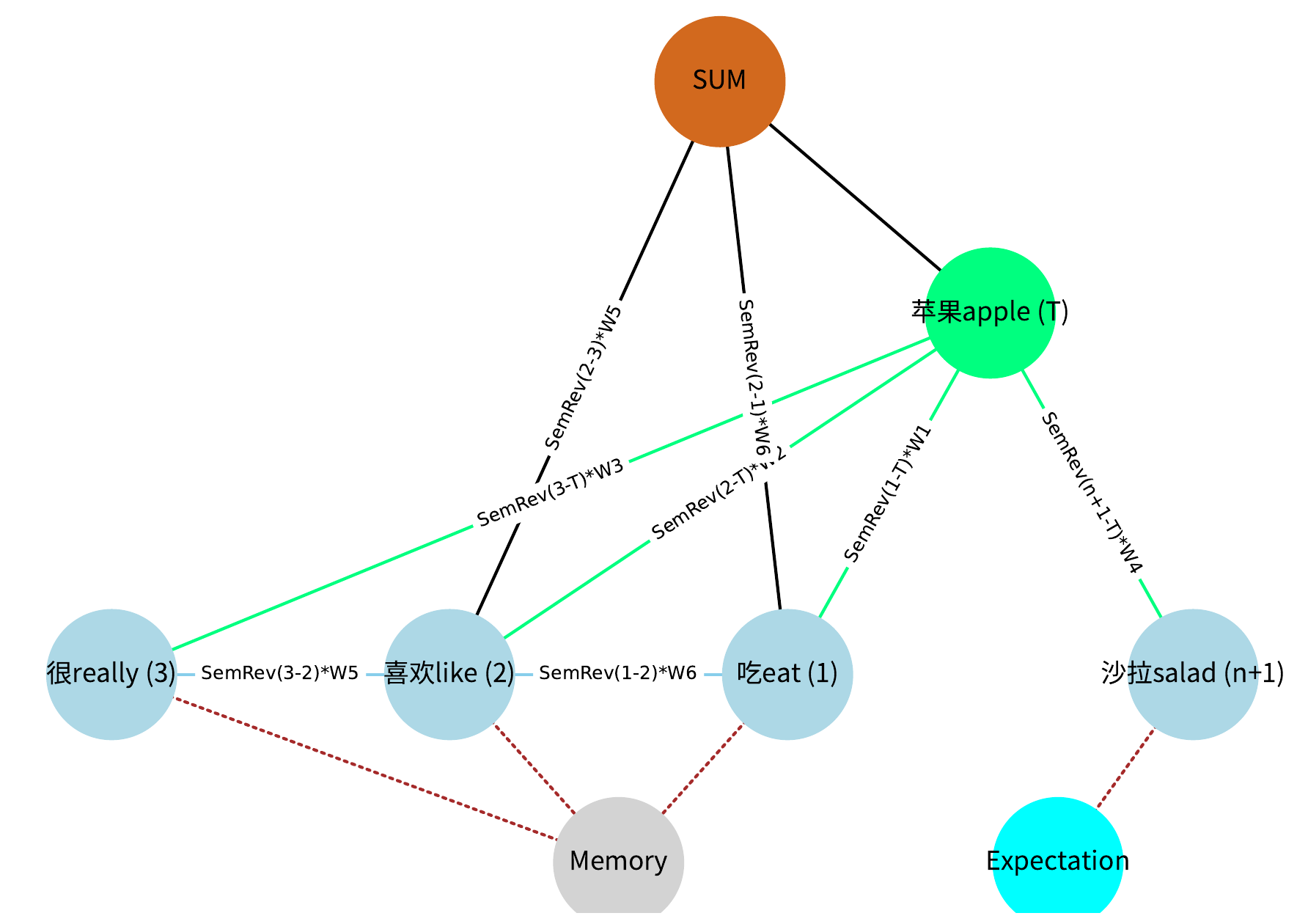}
		\caption*{\footnotesize{\textbf{B}.The ``attention-aware'' approach}}
		\label{fig:imageB}
	\end{minipage}
\caption{The computation of attention-aware metrics. The Panel A represents attention types used in Transformer. The Panel B illutrates the computational details of attention-aware metrics. \scriptsize{(Note: The example in the Panel B is \begin{CJK*}{UTF8}{gbsn}``我很喜欢吃苹果沙拉。''\end{CJK*}. The translation of the sample sentence is ``I/really/like/eat/apple/salad (I really enjoy apple salads.)''). The window comprises  five words: ``really'', ``like'', ``eat'', ``apple'', and ``salad'', with ``apple'' as the target word (T). Each word is assigned an ordinal number based on its position relative to the target word. ``SemRev'' denotes the semantic relevance between two words, which can be calculated using \textit{cosine} similarity or \textit{correlation}. For example, ``SemRev(2-T)'' indicates the semantic relevance between the second word (``like'') and the target word (``apple''). \textit{W} represents a weight ranging from 0 to 1, determined by the proximity of a contextual word to the target word. The expression ``$SemRev(2-T)*W2$'' quantifies the specific contribution of ``like'' to the realistic contextual influence on ``apple''. Six such values are aggregated at the node ``SUM'', signifying their summation. The three preceding words reflect memory information, while the subsequent word ``salad'' conveys expectation information; all are integrated into ``SUM''. The ``SUM'' value reflects the degree of semantic connection of ``apple'' within its context. )}}
	\label{fig:attention}
\end{figure}

	
	The Panel B of Fig. \ref{fig:attention} provides a specific example that illustrates the computational details. The \textit{dynamic} approach considered the preceding three words as a window (e.g., ``really'', ``like'', ``eat'', here we use English translations to represent corresponding Chinese words for convenience). In this window, any of the three words is computed as being semantically related with the target word (``apple''), which is shown in Fig \ref{fig:attention}. However, we can expand the window size by including the following word (e.g., ``salad''), which acts as an expectation for the target word. By adding the semantic relevance \footnote{Either \textit{cosine} similarities or \textit{correlations} among word embeddings serve as a computational method to calculate the semantic relevance between two words. In contrast, prior research employed correlation analysis to investigate the relationship between word vectors and reading duration, where correlation was a statistical method used to measure the association between two variables.} between the expected word and the target word to the value computed by the dynamic method, we obtain a value that represents how the target word is semantically related to the context, i.e., its contextual semantic relevance. However, this approach does not take into account the varying contributions of the surrounding words and the information of word order. To address this, we consider the different positional distance between a contextual word and the target word as a weight to compute its contribution. For instance, the weight of the relevance between ``really'' and ``apple''  would be lower since ``really'' is far from ``apple'', whereas the weight of the relevance between ``eat'' and ``apple'' would be higher since ``eat'' is in close proximity to ``apple''. This weight algorithm also takes the word order into account and can comprehensively incorporate contextual information.

The following details how to compute these metrics using the example in Fig. \ref{fig:attention}. In this window, there are five words,  including the target word ``apple'', which is preceded by ``really'', ``like'', and ``eat'' and followed by ``salad''. Each word is assigned a word vector from the pre-trained embedding database. First, to compute the \textit{dynamic} approach \citep{sun2023interpretable}, we calculate three semantic relevance values between the target word and each of its three preceding words, and two semantic relevance values between any two of the three preceding words. We then add these five semantic relevance values to get a summarized value. This method did not initially take into account the expected word, so we name it as ``\textit{contextual semantic relevance (- expectation)}''. Second, we calculate ``attention-aware'' semantic relevance. The upcoming word that is about to appear (e.g., ``salad'') creates an expectation for the target word (e.g., ``apple''). Put it simply, the word ``salad'' that follows the target word ``apple'' is not yet visible to the reader when they first encounter ``apple''. 
In this sense, after including the semantic relevance between the target word and the expected word (``salad''), we define the metric as ``\textit{attention-aware semantic relevance (- weights)}''. Now we need to include weights for each semantic relevance value. There are six semantic relevance values, and each  of them is multiplied by different weights. Weights range between 0 and 1 and are determined by the distance between the target word and contextual words. When a contextual word is closer to the target word, it is assigned a higher weight, and vice versa. For example, ``SemRev$_{T-3}$''(i.e, the semantic relevance between ``apple'' and ''really'') is multiplied by 1/2, and ``SemRev$_{T-2}$'' (the semantic relevance between ``apple'' and ``like'') is multiplied by 2/3, and ``SemRev$_{T-1}$'' (semantic relevance between ``apple'' and ``eat'') is multiplied by 1. The semantic relevance value for ``really'' and ``like'' is also reduced as they are not directly adjacent to the target word.  For instance, ``SemRev$_{2-3}$'' (semantic relevance between ``really'' and ``like'') is multiplied by 1/3, and ``SemRev$_{1-2}$'' (semantic relevance between ``like'' and ``eat'') is multiplied by 1/2. Similarly, ``SemRev$_{t-n1}$'' (semantic relevance between ``apple'' and ``salad'') will be reduced by a factor of 1/3 since we believe the expectation effect is not as strong as memory-based information.  After each semantic relevance value is multiplied by its corresponding weight, we add up all the values, and the final value is the ``\textit{attention-aware semantic relevance (+ weights)}''. \textbf{The \textit{attention-aware semantic relevance} represents the degree of semantic relatedness between the target word and the context}. In the example illustrated in Fig. \ref{fig:attention}, the \textit{attention-aware semantic relevance} of ``apple'' indicates the extent of its semantic association with the context. 

 The computation of attention-aware semantic relevance could be formalized as: \( \boxed{\sum SemRev_{(T,\ x)} \cdot W_{(T,\ x)}} \) (here \textit{T} = target word, \textit{x} = surrounding words, \textit{SemRev} = semantic relevance, \textit{W} = weights).  A higher value of attention-aware semantic relevance indicates a strong semantic relatedness of the target word with the context, and vice versa. Attention-aware semantic relevance serves as a computational metric that accurately assesses the contextual semantic degree, offering advantages over traditional methods reliant on human-rated semantic plausibility. It is increasingly employed in relevant studies as a more efficient and precise alternative.

 After detailing the computation, we focus on the cognitive interpretability of the new approach. The preceding words in this window illustrated in Fig. \ref{fig:attention} simulate a memory stack because readers have their memory of previously encountered words and their meanings. The weights in the attention-aware approach mimic humam forgetting and attentional allocation mechanisms (see in the Panel B of Fig. \ref{fig:mem}). 
 	Essentially, ``attention-aware'' signifies the ability to discern and function as efficiently as the ``attention'' algorithm in Transformer in capturing contextual information. That is why the approach is termed. 
 		From a cognitive standpoint, the ``attention-aware'' approach we proposed offers an interpretable means of computing semantic relevance from the context to the target word. The ``attention'' in the ``attention-aware'' approach should not be confused with human cognitive attention, despite our method considers the factors of human attention and memory processes. For instance, attention often refers to the cognitive mechanism by which individuals selectively focus on specific information or stimuli. Some studies posit that human attention during reading resembles a spotlight moving from word to word \citep{lewisa2005activation}. In contrast, the ``attention-aware'' approach asserts that contextual information significantly influences the reading process from one word to the next. As the reader progresses from one word to another, the processing is informed by preceding contextual information and the reader’s memory of the context. Despite this, our approach assigns weights to the semantic relevance between word pairs in a window, effectively mapping the distribution of attention during reading. Overall, the attention-aware approach can simulate the dynamic process of memory and attention allocation during reading to a certain extent.  
 
 The weight values decrease in terms of the distance between the target word and the contextual word, and weight plays a similar role in the human forgetting curve \citep{murre2015replication}. The attention-aware approach is mainly memory-based and has implications related to forgetting and memory. The weights utilized to compute attention-aware metrics are determined by the proximity between the target word and surrounding words. When the distance is short, a higher weight is assigned, and vice versa. In the current study, we adopted a window size of 4-5 words with varying weights mimicking the decline of memory retention over time. The forgetting curve depicts retained information halving after each day within a span of several days \citep{murre2015replication}. Specifically, in the current study, we used a five word window size to approximate the forgetting curve—a graphical representation of memory retention's decline over time, where information retention halves daily over several days. To mimic human memory decay, we assigned higher weights to nearer words and lower weights to more distant ones. This algorithm parallels the human forgetting pattern with the attentional weights in our study, as shown in the Panel B of Fig. \ref{fig:mem}, effectively capturing the influence of contextual words and their order. The inclusion of both the expected word and weights stems from the consideration of the expectation effect and memory mechanisms. Without these features, semantic relevance only considers memory factors. In the discussion section, we will provide a detailed explanation about this, and delve into the underlying mechanisms that contribute to their memory-based nature and how they differ from other types of metrics.

	In summary, the proposed method for calculating attention-aware metrics involves incorporating the expectation and weight factors. This allows for a larger window size to be considered, including both preceding and the following word as potential sources of contextual information. The relevance values between these contextual words and the target word are then weighted based on their position distance and added together to obtain a comprehensive metric. Moreover, as mentioned in section 2.3, the use of attention and memory, represented as weights, is important in memory-based information. Weights help to focus on relevant information while ignoring irrelevant information, which enhances the ability to retrieve memories accurately and efficiently. 
	
Additionally, to facilitate a comprehensive comparison, we computed ``word surprisal'' for the stimuli text for the corpus of eye-tracking data on Chinese reading \citep{zhang2022database}. The two state-of-the-art large language models (LLMs) were taken to estimate word surprisal: Chinese BERT \citep{cui2021pre} and multilingual GPT \citep{shliazhko2022mgpt}. We can leverage pre-trained LLMs to estimate word probabilities and subsequently obtain word surprisals. In BERT, some words in the input sentences are randomly masked or replaced with a special [MASK] token. BERT's ability to predict masked words and capture contextual information allows for computing word surprisal. GPT is an autoregressive language model, which means it generates text one word at a time while considering the context of previously generated words. When predicting the next word in a sequence, GPT takes into account all the preceding words in the sentence. This allows GPT to compute word surprisal. The expectation-based metric (i.e. surprsial) could be valuable in contrasting with the semantic-based metrics proposed in the present study. Overall, this study used the four methods (i.e.,\textit{surprisal}, \textit{cosine}, \textit{dynamic}, and \textit{attention-aware}) to yield two types of word surprisal, and four types of semantic-based metrics, as summarized in Table \ref{table:semtype} \footnote{The code for implementing the attetnion-aware approach and computing surprisal, and a data sample available at: \url{https://osf.io/gd7w8/}}.

	\subsection{Statistical methods}

 
To address the two research questions, which concern whether these metrics can accurately predict eye-tracking data and which metrics perform better, we employed Generalized Additive Mixed Models (GAMMs) \citep{wood2017generalized}. GAMMs are a type of mixed-effects regression model that are particularly useful for analyzing (non)linear effects and multiplicative interactions between variables, making them well-suited for assessing the predictability of these metrics. Compared to traditional regression methods, GAMMs offer greater flexibility in modeling complex relationships between variables.
	
	The use of statistical methods is more straightforward when response variables are simple. Eye-tracking data has the advantage of being easier to analyze with these methods compared to EEG and fMRI data, making it the ideal choice for our study on Chinese naturalistic discourse reading. However, assessing model performance and comparing models can be complex, and it can be challenging to gain insights into the predictability of the metrics through \textit{correlations} alone. Fortunately, GAMMs are well-suited for comprehensive and precise assessment of model performance. In our study, we compared the models using \texttt{AIC} (Akaike Information Criterion), with the criterion being that a smaller value indicates a better model. 
	
	Regarding model comparison, we compute the difference in $\Delta$\texttt{AIC} values for each GAMM  by subtracting the \texttt{AIC} of the base GAMM model from the \texttt{AIC} of a full GAMM model. The base model did not incorporate the independent variable of our interest, but the full model included it.  
	These allow us to assess the relative fit of different models and aid in selecting the most appropriate model for our analysis. Note that the word frequency was provided by the dataset of Chinese reading.
	Table \ref{table:semtype} provides an overview of the metrics and analysis methods.
	


	\begin{table}[h!]
		\centering
		\caption{The methods and metrics used in this study}
		\scalebox{0.6}{
			\begin{tabular}{c c c c} 
				\toprule
				\textit{Method} & \textit{Metric} & \textit{Algorithm} & \textit{Regression Analysis} \\ 
				\midrule
				surprisal & \makecell{word surprisal computed\\ by BERT and GPT respectively} & \makecell{\scriptsize word probability computed by neural masked\\ \scriptsize language modeling in BERT,\\ \scriptsize and language modeling in GPT.} & GAMM \\
				\addlinespace
				cosine & \makecell{cosine semantic similarity\\ (same as \citet{frank2017word}\\ but consider both function and content words)} & \makecell{\scriptsize The cosine value between the sum of the vectors\\ \scriptsize of the preceding three words and the vector of the target word} & GAMM \\
				\addlinespace
				dynamic & \makecell{attention-aware semantic relevance (- expectation)\\ (i.e.,dynamic semantic similarity \citep{sun2023interpretable})} & \makecell{\scriptsize The sum of the semantic relevance values between\\ \scriptsize the target word and the three preceding words is calculated} & GAMM \\
				\addlinespace
				attention & \makecell{attention-aware\\ semantic relevance (- weights)} & \scriptsize the information of the next word (n+1) is included, but no weights are considered & GAMM \\
				\addlinespace
				attention & \makecell{attention-aware\\ semantic relevance (+ weights)} & \scriptsize the information of weights and n+1 word is included & GAMM \\
				\bottomrule
			\end{tabular}
		}
		\label{table:semtype}
	\end{table}

	\section{Results}


	\subsection{Study 1: GAMMs with random slope}
	

	    First of all, we did correlation analysis on all metrics of our interest, and found that the two types of surpisal is less correlated with semantic-related metrics, shown in Fig. \ref{fig:corr} (see the Appendix). It indicates that our attention-aware metrics are distinct ones from word surprisal. In contrast, the several semantic-based metrics are highly correlated.
	     
		Second, the past studies show that a number of factors have an impact on fixation durations on word reading. Some factors, such as \texttt{word frequency}, \texttt{word length} have been extensively investigated to show their significant effects on reading, and these factors could be taken as control predictors in reading. However, as for Chinese reading, we believe that ``word stroke count'' is taken as a control predictor better than ``word length''. Using the eye-movement measures on words in Chinese reading \citep{zhang2022database}, we applied GAMM fittings to compare which factor performs better in impacting on Chinese word reading, and the result shows that `` word stroke count'' remarkably outperformed ``word length''. The present study took ``word stroke count'' as a control predictor in GAMM fittings. The result and analysis is seen in the Appendix. 
		
		 Third, we fitted 18 GAMM models to analyze word surprisal and the four types of semantic relevance as main predictors of interest for three dependent variables \textit{total duration}, \textit{first duration}, and \textit{gaze duration}. The models also include \textit{stroke count of word} and \textit{word frequency} as control predictors, modeled as tensor interaction, and \textit{experiment} as a random effect. The dataset of eye-movement does not provide the information on ``participant'', so we used ``experiment'' as a random variable.) The main predictor of interest is modeled as a tensor product smooth. A GAMM model looks like this: \texttt{log(duration) $\sim$ te (stroke\_count, log\_freq) + s(metric) + s (experiment, bs=``re'')} (here, \texttt{s} = tensor product smooth, \texttt{te} = tensor interaction, \texttt{re}=random effect). Note that the response variable was log-transformed to approximate a normal distribution more closely, thereby achieving better model fit. 
		

	The following first reports effects of word surprisal on Chinese reading. When the random variable \texttt{experiment} was included the GAMM fittings, two types of word surprisal—calculated using BERT and GPT—showed no significant effect on fixation durations (with a \textit{p} -value much greater than 0.05). Even if a GAMM fitting only included \texttt{word surprisal} with the random effect of \texttt{experiment}, \texttt{word surprisal} was not significant either. Nonetheless, incorporating random effects is crucial in the analysis of fixed-effect statistical models \citep{baayen2008mixed}. Note that the \texttt{word surprisal} metric was transformed logarithmically to approximate a normal distribution, enhancing model fitting. These results indicate that \textbf{word surprisal may have an weak effect on eye-movements during Chinese reading}. 

\begin{figure}[!h]
		\centering
		\includegraphics[width=0.98\textwidth]{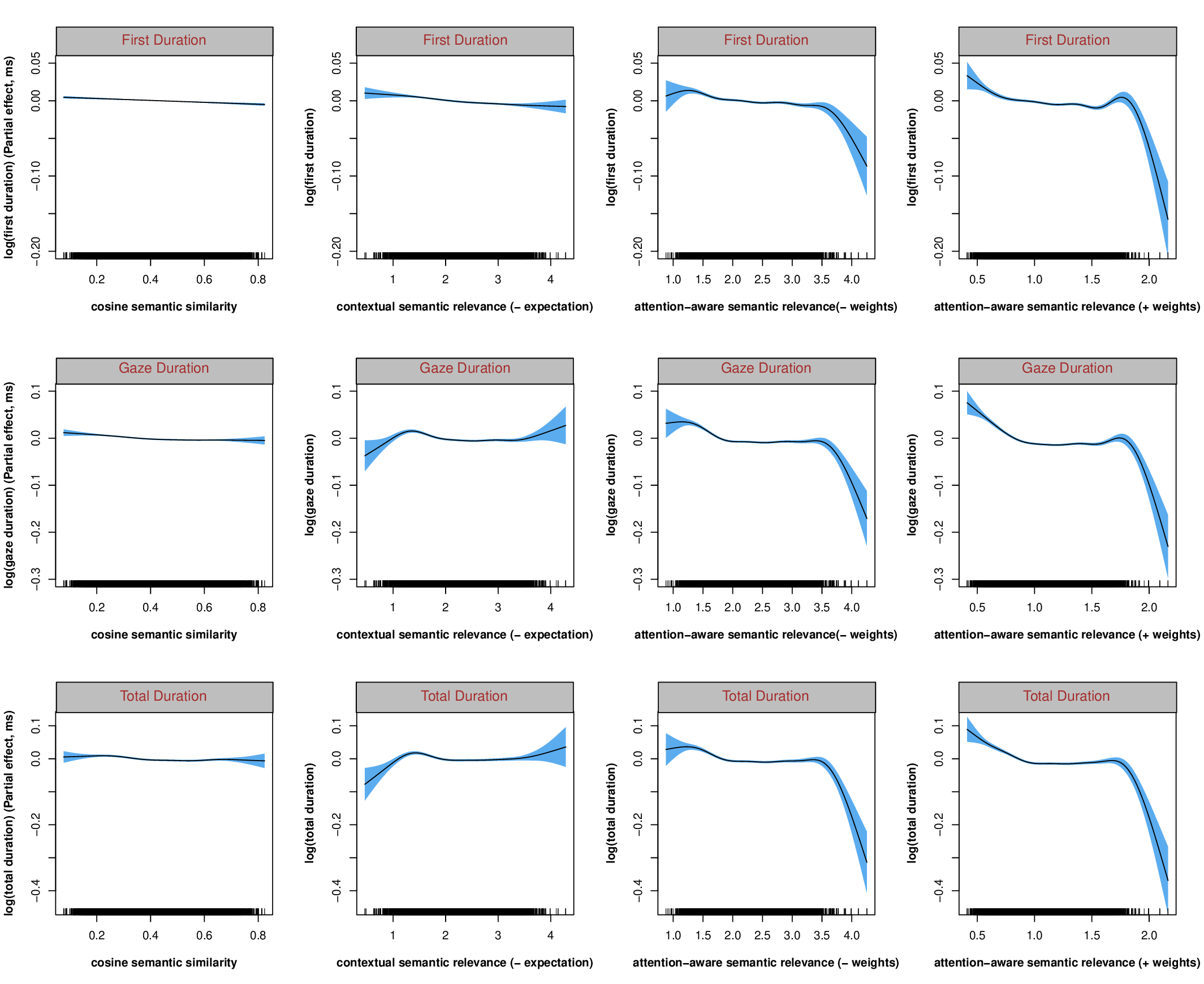}
		\caption{Partial effects for various metrics of semantic relevance. \footnotesize{The \textit{x}-axis represents the metric, while the y-axis signifies the fixation duration. Each curve in one plot demonstrates the relationship between a predictor variable and the response variable. More pronounced slopes on these curves signify a more substantial influence of the predictor on fixation durations, whereas more gentle slopes imply a less significant effect. ( Note: n = 76549; both \textit{attention-aware semantic relevance (- weights)} and \textit{attention-aware semantic relevance (+ weights)} are \textit{attention-aware} metrics; \textit{contextual semantic relevance (- expectation)} is not a typical attention-aware metric. ``log'' signifies log-transformation.)} }
		\label{fig:partial effects}
	\end{figure}

	  We then report the results of semantic-based metrics effects. In Fig. \ref{fig:partial effects}, the partial effects of the semantic relevance variables on first duration, gaze duration and total duration in the GAMM models are presented. It is evident that all significant effects have a negative influence, which means that fixation duration tends to decrease as semantic relevance increases. These findings suggest that Chinese users require more time to process words that have lower semantic relevance, i.e., words that are less likely to occur in the context. On the other hand, words that have higher semantic relevance require less time to process as they are more likely to occur in the same context. 
	 However, different types of semantic relevance take effect at different stages. For instance, the \textit{cosine similarity} is activated earlier, but its effect disappears quickly. In contrast, the \textit{contextual semantic relevance (- expectation)} has a stronger effect on reading durations in the three cases, compared to the \textit{cosine similarity}. However, compared to \textit{contextual semantic relevance (- expectation)}, the two \textit{attention-aware} metrics have a much stronger and more stable effect on the three types of duration. These differences may have arisen because of the diverse approaches used to obtain semantic information. 
	 
The results presented above indicate that \textit{cosine similarity} is not as effective in predicting fixation duration compared to the other three types of semantic relevance. As illustrated in Fig. \ref{fig:partial effects}, the \textit{cosine similarity} curves remain relatively flat across all three cases. By examining the curve trends in Fig. \ref{fig:partial effects}, we can identify which measure produces a stronger effect on reading duration. We further compared the performance of the models by using $\Delta$\texttt{AIC}.  
Specifically, the basis for comparison is the consistent data point numbers (n = 76727) and identical elements in each GAMM fitting. $\Delta$\texttt{AIC} values between GAMM by subtracting the \texttt{AIC} of the base GAMM model from the \texttt{AIC} of a full GAMM model. The base GAMM model did not include the independent variable we are interested. The comparison of $\Delta$\texttt{AIC} values plays a crucial role in evaluating the performance of the GAMM fitting. A smaller $\Delta$\texttt{AIC} value suggests a better fit for the GAMM fitting, indicating superior performance for the metric used. In simpler terms, a smaller $\Delta$\texttt{AIC} signifies that the model performs more effectively and aligns more closely with the observed data. Thus, we can rely on the $\Delta$AIC values to guide us in selecting the most suitable GAMM for our analysis. Table \ref{tab:compare} provides the data on $\Delta$\texttt{AIC} for comparing the GAMM fittings.

\begin{table}
	\centering
	\caption{$\Delta$\texttt{AIC} for GAMM fittings with different metrics using random slope (n = 76727). ``log'' = log-transformation}
	\scalebox{0.7}{
		\begin{tabular}{l *{3}{c}}
			\toprule
			\textbf{GAMM fittings with semantic metrics} & {\textbf{log(First Duration)}} & {\textbf{log(Gaze Duration)}} & {\textbf{log(Total Duration)}} \\
			\midrule
			cosine semantic similarity             & -31    & -77.6  & -68.58 \\
			contextual semantic relevance (- expectation)   & -84.6  & -139.9 & -79.8  \\
			\textcolor{teal}{attention-aware semantic relevance (- weights)}    & \textcolor{teal}{-119}  & \textcolor{teal}{-304.4} & \textcolor{teal}{-188.09} \\
			\textcolor{blue}{attention-aware semantic relevance (+ weights)}  & \textcolor{blue}{-167.2} & \textcolor{blue}{-479.1} & \textcolor{blue}{-297.82} \\
			\bottomrule
		\end{tabular}
	}
	\label{tab:compare}
\end{table}

Table \ref{tab:compare} provides the $\Delta$\texttt{AIC} data for all GAMM fittings, with consistent datapoint numbers (n = 76727) and identical elements in each model. When the $\Delta$\texttt{AIC} value is negative, it indicates that the inclusion of this metric in the GAMM fitting improves the model's performance compared to the base GAMM model. In other words, the negative value suggests that the metric contributes positively to the model's fit, leading to a better representation of the observed data. Further, a smaller $\Delta$\texttt{AIC} value indicates better model performance, and vice versa. According to Table \ref{tab:compare}, these metrics contribute positively to GAMM fittings, that is, these metrics show their predictability in eye-movement on Chines reading. For \textit{Total Duration}, the $\Delta$\texttt{AIC} of \textit{attention-aware semantic relevance (+ weights)} is 230 smaller than that of \textit{cosine similarity}, 219 smaller than that of \textit{contextual semantic relevance (- expectation)}, and 110 smaller than that of \textit{attention-aware semantic relevance (- weights)}. The \textit{attention-aware semantic relevance (+ weights)} consistently has the smallest $\Delta$\texttt{AIC}, while \textit{attention-aware semantic relevance (- weights)} has the second smallest. In contrast, the $\Delta$\texttt{AIC} of \textit{cosine semantic similarity} is the largest. These data suggest that GAMM models incorporating \textit{attention-aware semantic relevance (+ weights)} demonstrate the best performance.


  	\subsection{Study 2: GAMMs with random smooth}

   We then fitted another group of 18 GAMM fittings with random smooth to analyze the six metrics as predictors of three dependent variables from the same Chinese eyemovement dataset. The main predictor of our interest is modeled as a tensor product smooth. The random effect is also the same as in Study 1. The GAMM fittings also include \textit{word stroke count} and \textit{word frequency} as control predictors, modeled as tensor interaction, and \textit{experiment} as a ``random smooth''. An optimal GAMM fitting is formulated as: \texttt{log(duration) $\sim$ te (word\_length, log\_wordfreq) + s(metric, experiment, bs =``fs'', m = 1), data = data} ( \texttt{s} = tensor product smooth, \texttt{te} = tensor interaction, \texttt{fs} = random smooths adjust the trend of a numeric predictor in a nonlinear way, and it cover the function of random intercept and random slope; the argument \texttt{m=1} sets a heavier penalty for the smooth moving away from 0, causing shrinkage to the mean). In Study 1, ``experiment'' was taken as random effect (i.e., ``re''), and random effect is random slope adjusting the slope of the trend of a numeric predictor.  In contrast,  in this GAMM equation, \texttt{experiment} is treated  as random smooth. Random smooth leverages random slope and random intercept to fully
   assess the significance of the metrics of interest. In other words, random smooth could examine random effect more comprehensively, including both random slope and random intercept. 
   
   Some variables were log-transformed in order to make the data closer to normal distribution, and thus achieving better fittings. First of all, \texttt{word surprisal} did not show significant results (indicated by a \textit{p}-value exceeding 0.05). Compared to the GAMM fittings that incorporate a random slope, the introduction of a random smooth for \texttt{experiment} did not change \texttt{AIC} value. These findings reveal that neither variant of word surprisal (calculated via BERT or GPT) had significant effect on any of the three examined fixation durations. This aligns with earlier results obtained from models including random slope.
   
   We then used a higher threshold of \textit{p}-value $<$ 0.01 to determine the significance of variables in a GAMM fitting. The results of GAMM fittings with random smooth show that all metrics have an effect on the three types of durations quite well. Despite this, attention-aware metrics have stronger predictability in the eye-movement data. As shown in Table \ref{tab:compare1}, attention-aware semantic relevance with weights has the smallest \texttt{$\Delta$AIC}, indicating that the metric has the strongest effect on the three eye-movement data. In addition, the tensor interaction between \textit{word length} and \textit{word frequency} has a significant effect, suggesting that both control predictors could remarkably predict eye-movements on reading. Additionally, the random effect of the \textit{participant} is strongly significant in all GAMM fittings. The results of GAMM fitting using random smooth are consistent with those using random slope. 
   
 \begin{table}
 	\centering
 	\caption{$\Delta$\texttt{AIC} for GAMM fittings with random smooth (n = 76549) \footnotesize{(Note: The value enclosed in square parentheses ($[\quad]$) within each cell indicates the difference in \texttt{AIC} between the GAMM fitted with a random slope (see Table \ref{tab:compare}) and the model fitted with a random smooth. Both GAMM fittings incorporate identical elements. A negative value suggests that the metric of interest contributes more significantly to the model with the random smooth than to the one with the random slope. A smaller value signifies a greater contribution of the metric to the GAMM fitting. In essence, a lower value within square parentheses denotes superior performance of the metric.)}}
 	\scalebox{0.7}{
 		\begin{tabular}{l *{3}{c}}
 			\toprule
 			\textbf{GAMM fittings with semantic metrics} & {\textbf{log(First Duration)}} & {\textbf{log(Gaze Duration)}} & {\textbf{log(Total Duration)}} \\
 			\midrule
 			cosine semantic similarity & -130.2 [-98.8] & -246.3 [-147.7] & -260.84 [-188.44] \\
 			contextual semantic relevance (- expectation) & -536.9 [-452.3] & -681.6 [-514.9] & -929.38 [-814.78] \\
 			\textcolor{teal}{attention-aware semantic relevance (- weights)} & \textcolor{teal}{-934 [-827]} & \textcolor{teal}{-1410 [-1062.5]} & \textcolor{teal}{-1692.12 [-1480.22]} \\
 			\textcolor{blue}{attention-aware semantic relevance (+ weights)} & \textcolor{blue}{-1163.7 [-1030.9]} & \textcolor{blue}{-1798.5 [-1271.9]} & \textcolor{blue}{-2225.84 [-1862.74]} \\
 			\bottomrule
 		\end{tabular}
 	}
 	\label{tab:compare1}
 \end{table}
 
	These comparisons using partial effect curves and $\Delta$\texttt{AIC} yield the same results, with the models with \textit{attention-aware semantic relevance} performing the best for three response variables. Specifically, for total duration, the performance of the models is ranked as follows: {\color{blue}{ \textit{attention-aware semantic relevance (+ weights) } $>$ \textit{attention-awarel semantic relevance (- weights) }  $>$ \textit{contextual semantic relevance (- expectation) }  $>$ \textit{ cosine similarity}}}; for first duration: {\color{blue}{ \textit{attention-aware semantic relevance (+ weights) } $>$ \textit{attention-aware semantic relevance (- weights) }  $>$ \textit{contextual semantic relevance (- expectation) }  $>$ \textit{ cosine similarity}}}; regarding gaze duration: {\color{blue}{ \textit{attention-aware semantic relevance (+ weights) } $>$ \textit{attention-aware semantic relevance (- weights) }  $>$ \textit{contextual semantic relevance (- expectation) }  $>$ \textit{ cosine similarity}}}.

In short, the \textit{dynamic} contextual semantic relevance has a better performance than the \textit{cosine} similarity. However, both attention-aware metrics outperform either \textit{dynamic} one or \textit{cosine} one.  All this indicates that attention-aware metrics are highly capable of predicting different types of eye-movements on Chinese naturalistic reading.

	\section{Discussion}
	
	
	To recap, our GAMM analyses produced consistent results. 
	First, the effect of word surprisal on Chinese reading took place weakly, which is different from English and other languages. Second, the two \textit{attention-aware} metrics were the best predictors. Third, the \textit{cosine} metric also predicted all three response variables, although the \textit{dynamic} semantic relevance outperformed it. Overall, the \textit{attention-aware} metrics were superior to other metrics in predicting all three response variables. The following gives a further analysis of our findings. 
	


\subsection{Memory-based advantages and parallel processing}
	
This section delves into the effectiveness of attention-aware metrics in predicting Chinese reading. The notable success of these metrics stems from their exceptional ability to capture contextual information and emulate human reading processes. The underlying strength of the attention-aware approach mainly lies in its functionality as a memory processor, adept at storing, retrieving, and integrating information. 

\begin{figure}[!h]
	\centering
	\includegraphics[width=0.98\textwidth]{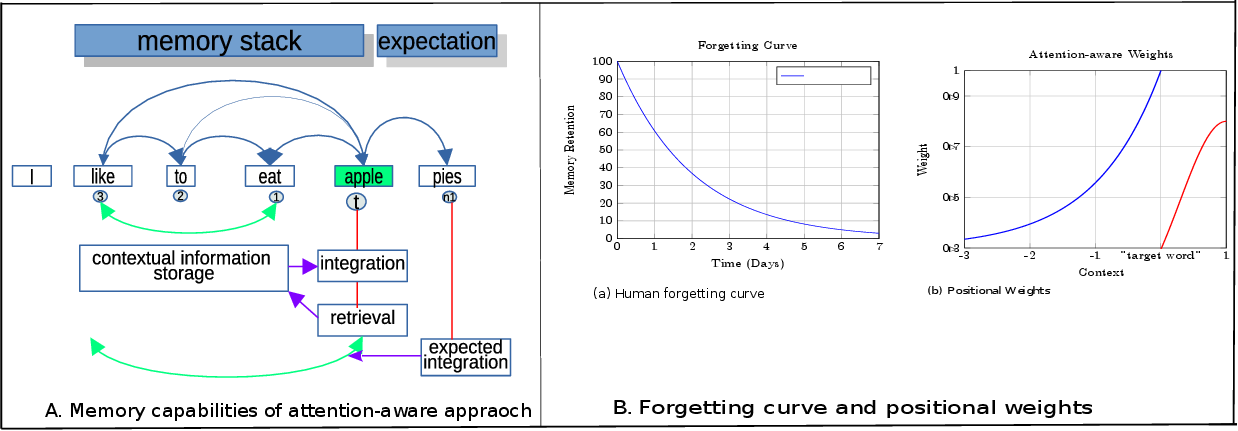}
	\caption{The memory capability of the attention-aware approach and positional weights}
	\label{fig:mem}
\end{figure}

Attention-aware metrics are mainly considered memory-based. As mentioned earlier, the preceding words in a window illustrated in Fig. \ref{fig:attention} simulate a memory stack because readers have their memory of previously encountered words and their meanings, as shown in the Panel A of Fig. \ref{fig:mem}. The information of the preceding words is stored in such a stack by integrating it with the target word, while the subsequent word is similarly integrated. Contextual information can be retrieved by connecting it with the target word in the stack. This stack moves forward in a sliding manner, working as a sliding memory processor. In this way, the attention-aware approach facilitates the storage and decay of contextual information, acting as a memory agent/processor to capture and manage the contextual information effectively. Moreover, when utilizing a 4-word window, which is similar to the phenomenon of memories fading within four days, the rate of memory fading can be approximated at an average of one-quarter per day. However, this fading process occurs more rapidly at the outset and decelerates over time. Following this model, we can adjust the significance of words relative to their proximity to the target word by reducing their importance by a quarter. Consequently, words more distant from the central word are assigned diminished weights compared to a uniform distribution. This results in a gradation of weights, such as 1/3, 2/3, 1/2, etc., reflecting their varying degrees of closeness to the target word. 

\citet{van2022explaining} proposed an accurate model requiring a forgetting function that changes with time intervals. In the attention-aware approach, assigning different weights based on the distance between the target word and surrounding words \footnote{In the future, we plan to incorporate factors such as word dependency distance \citep{liu2017dependency}, the distinction between content and function words, among others, to refine the weighting system. This adjustment will make the weights more informative, thereby enhancing the precision of the metrics in their predictive capability.} is similar to how time intervals influence the forgetting function in memory models. The word proximity signifies a measure of semantic or contextual difference. Words that are closer to the target word receive a larger weight, reflecting their greater relevance and potential impact on comprehension and memory. The reason for this is that words in close proximity to the target word indicate initial phases of memory decline, whereas those at a greater distance signify advanced stages, as shown in the Panel B of Fig.\ref{fig:mem}.  Further, in memory models such as the ACT-R model\citep{anderson1997actr}, the word proximity can be likened to the time interval between learning and recall. Our attention-aware processor has the similar mechanism: shorter intervals correspond to better recall (higher ``weight''), while longer intervals lead to more forgetting (lower ``weight''), shown in the Panel B of Fig. \ref{fig:mem}.

Our findings support this point. The attention-aware metric without using weights (i.e., attention-aware semantic relevance (- weights)) did not outperform the metric using weights (i.e., attention-aware semantic relevance (+ weights)) in predicting eye-movement on Chinese reading, as shown in Table \ref{tab:compare1}. The metric without using weights does not account for the forgetting/memory situation, and the metric did not perform as well as the metric considering weights. This finding suggests that the metric considering the human forgetting/memory feature exhibits stronger predictability than the one that overlooks this factor. Moreover, our finding indicates the existence of a forgetting effect in reading, and a potential connection between the attention-aware approach and the forgetting curve. Both involve a form of ``decay'' or decrease as the time distance (interval) increases.

Next we further discuss how our metrics assess E-Z reader and SWIFT model (\citealp{pollatsek2006tests}; \citealp{engbert2005swift}). The E-Z Reader model refutes the advantage of previewing subsequent words, suggesting that reading progresses linearly/serially. Conversely, the SWIFT model recognizes the benefits of previewing upcoming words and maintains that language processing occurs in parallel. Our attention-aware approach facilitates the simulation of both E-Z reader and SWIFT model with ease. Our attention-aware approach created three forms of semantic relevance metrics, with the first one (i.e, the metric without expectation) resembling the E-Z reader, while the latter two metrics are more akin to the SWIFT which supports preview benefits. Our results show that the attention-aware metric without incorporating the next word information did not outperform the metrics considering the next word information (in some studies, that is ``n+1'' word), shown in Table \ref{tab:compare1}. This suggests that semantic preview benefits could occur in reading Chinese, and the semantic preview effects could be proven in a massive dataset of eye-movements on Chinese reading. Despite this, the attention-aware metric without the next word information is still a strong predictor, and this reveals that E-Z model can also explain reading behavior quite well. Morevoer, surprisal only uses the left context of the current word and does not consider the information from the ``n+1'' word, making it suitable for explaining serial processing. However, the surprisal effect could weakly occur in Chinese naturalistic reading.

Previously, carefully controlled sentence reading studies typically did not provide support for preview benefit effect. However, realistic reading scenarios, similar to naturalistic discourse reading, differ from lab-controlled sentence reading (\citealp{kennedy2005parafoveal}; \citealp{schotter2012parafoveal}). Our study's findings reveal that attention-aware metrics without next-word information do elicit effects, supporting the existence of serial processing in language comprehension/processing. Nonetheless, the heightened predictability of attention-aware metrics incorporating the next-word information suggests that parallel processing may take place during naturalistic reading without control. Conversely, under controlled conditions (e.g., reading controlled sentences in labs), serial processing might suppress parallel processing, potentially explaining the coexistence of both phenomena with the prominence of each contingent on specific conditions \citep{wen2019parallel} and language distinctions. The similar findings can be applied in reading other multiple languages \citep{sun2023attention}.

Our findings also confirm the effects of contextual semantic relevance on eye-movements in Chinese reading, as have reported in previous studies of the relationship between semantic similarity and eye-movements (or reading/comprehension difficulty) (e.g., \citealp{roland2012semantic}; \citealp{broderick2019semantic}; \citealp{sun2023interpretable}), and of the relationship between semantic plausibility and eye-movements (\citealp{hohenstein2010semantic}; \citealp{schotter2013synonyms};\citealp{yang2012semantic}; \citealp{li2023semantic}). Furthermore, the effect of word surprisal did not significantly manifest during naturalistic Chinese reading. The minimal influence of word surprisal on reading has been observed in certain languages (e.g., Korean, Finnish) \citep{sun2023optimizing}. Several factors may contribute to this. First, word surprisal, which indicates expectation, might not have a pronounced effect in Chinese comprehension. Second, Chinese readers might leverage the preview effect to mitigate expectation impacts during reading. Despite these findings, surprisal remains a valuable metric for reading in some languages like English. Third, Chinese, Korean, and Finnish belong to different language families, where users might employ other strategies to enhance comprehension more frequently. The long-standing debate over prediction or expectation continues (\citealp{huettig2016prediction}; \citealp{nieuwland2018large}; \citealp{poulton2022can}). The prediction effect may be influenced to some degree by a number of factors, such as lab conditions, experiment types, statistical analysis. However, considering language factors could offer clarity. That is, in certain languages, expectation strategies are used but not as dominantly, with a greater emphasis on memory strategies. Conversely, expectation strategies might be more heavily adopted in other languages. Thus, the surprisal or expectation-based model is not universally applicable across all languages for investigating language comprehension.

\subsection{Main methodological contributions}

There are several key methodological contributions in the current study. We introduced an attention-aware approach to compute contextual semantic relevance that surpassed existing approaches (\citealp{frank2017word}; \citealp{broderick2019semantic}; \citealp{sun2023interpretable}). Furthermore, the new approach is cognitively and linguistically interpretable, which is a significant advantage. 
Overall, the attention-aware approach considers both the semantic relatedness between words and weights of memory and attention, making it a powerful computational tool for predicting the processing difficulty experienced by readers, evaluating the different reading theories, and understanding the cognitive mechanisms underlying Chinese language processing.


 We believe that several reasons enable the attention-aware metrics to perform exceptionally well. First, contextual semantic relevance that incorporates both memory-based information and expectation-based information can better model how humans comprehend and process words in contexts. When reading or listening to language, language users not only rely on their memory of previously encountered words and their meanings (memory-based information), but also use their expectations of what may come next in the sentence or discourse (expectation-based information). 
 By considering the relationship between the current word and both the preceding and upcoming words, our approach can better capture the semantic relatedness between words and the context. 
 
 Second, by effectively incorporating different types of information, attention-aware semantic relevance may capture the complex interactions between attention, memory decay and semantic plausibility during language processing. 
 For example, when processing a sentence, a reader may allocate less attention to words that are more semantically related to the overall or contextual meaning of the sentence, while allocating more attention to words that are less relevant. Moreover, \textit{surprisal} is one of the most influential computational metrics in predicting language comprehension. However, \textit{surprisal} could not predict Chinese reading. By contrast, our approach provides a very different but effective method to predict Chinese reading. 

Third, 
the attention-aware approach we proposed offers convenient metrics for quantitatively evaluating various reading theories and language processing models, as discussed earlier. E-Z Reader vs. SWIFT model, as well as serial processing vs. parallel processing, have been validated through lab-based experiments and limited datsets. While only a limited number of computational metrics could be employed to test these models with massive datasets, our attention-aware approach is able to quantitatively assess these theories using extensive datasets and even conduct cross-language tests. For instance, the two models of eye-movement control in reading, E-Z Reader and SWIFT, assume that both word frequency (a memory-based metric) and cloze predictability (an expectation-based metric) play crucial roles in determining whether and how long words are fixated. However, the E-Z Reader and SWIFT model solely rely on word frequency and cloze predictability as separate metrics and lack the ability to generate their own metrics for evaluating and predicting eye-movements. In contrast, the attention-aware metrics we proposed can independently predict eye movements and also assess these reading models.

Finally, the attention-aware metrics can detect time-series data on language comprehension and processing with greater depth and precision. For example, our findings suggest that the attention-aware metrics can have an effect on both early and late stages of language processing, as shown in Fig. \ref{fig:partial effects} (see the partial effect (ms = milliseconds) of \textit{y}-axis in each plot). Some of our results align with those of previous studies. For instance, \citet{yan2020early} used the \textit{cosine} approach \citep{frank2017word} to compute semantic similarity and reported that contextual information affects ERPs in both early (200ms after word onset) and late (N400) time windows. Our results on \textit{cosine} metric are largely consistent with their findings. Similarly, \citet{broderick2019semantic} found that a word's semantic similarity to its sentential context enhanced the early cortical tracking of its speech envelope, which manifested in the prediction accuracy of EEG signals. We found similar results for \citet{broderick2019semantic}'s Euclidean semantic similarity in predicting eye-movements on reading. However, due to the weakness of the metrics they adopted, they could not find that the effect of contextual semantic relevance took place throughout the entire process of language comprehension. It is not the case that the effect of semantic information vanishes in language comprehension, but rather some methods may not capture semantic information effectively or precisely. Based on the evidence presented in Fig. \ref{fig:partial effects}, \textit{cosine} similarity is less effective in predicting processing difficulty as indicated by eye-tracking data. Additionally, it is insufficient for capturing the complete influence of semantic relevance during language processing in real-time experiments. While the effects at the early and late stages can be observed, detecting the effect at the intermediate stages remains challenging. On the contrary, attention-aware semantic relevance can detect the effects that occur throughout the entire word processing stage.
	
	\section{Conclusion}

 The present study introduced an attention-aware approach for computing contextual semantic relevance and used a database of eye-tracking data to investigate its effectiveness in predicting eye-movements during Chinese reading on naturalistic discourse. The results confirmed that both word stroke count and contextual semantic relevance computed by our approach were significant predictors of eye-movements in reading. In comparison to  the \textit{cosine} and \textit{dynamic} approaches, the metrics obtained through the attention-aware approach this study proposed exhibited superior performance. Our data provide evidence for semantic preview benefits and parallel processing in Chinese reading. However, they also demonstrate that the E-Z model and serial processing can also explain Chinese reading behavior. Although our focus was on Chinese eye-tracking data, the new approach can be applied to investigate eye-movement data in other languages, 
and data from EEG or fMRI experiments on language processing and comprehension, as well as visual processing. 
The attention-aware approach offers high interpretability and excellent predictabilityy, making it a valuable computational tool for modeling eye-movements in reading and language comprehension. 
 
	
	\bibliographystyle{apa}
	\bibliography{reference.bib} 
	

\appendix

\section{The Pearson correlation among the metrics}

\begin{figure}[!h]
	\centering
	\includegraphics[width=0.86\textwidth]{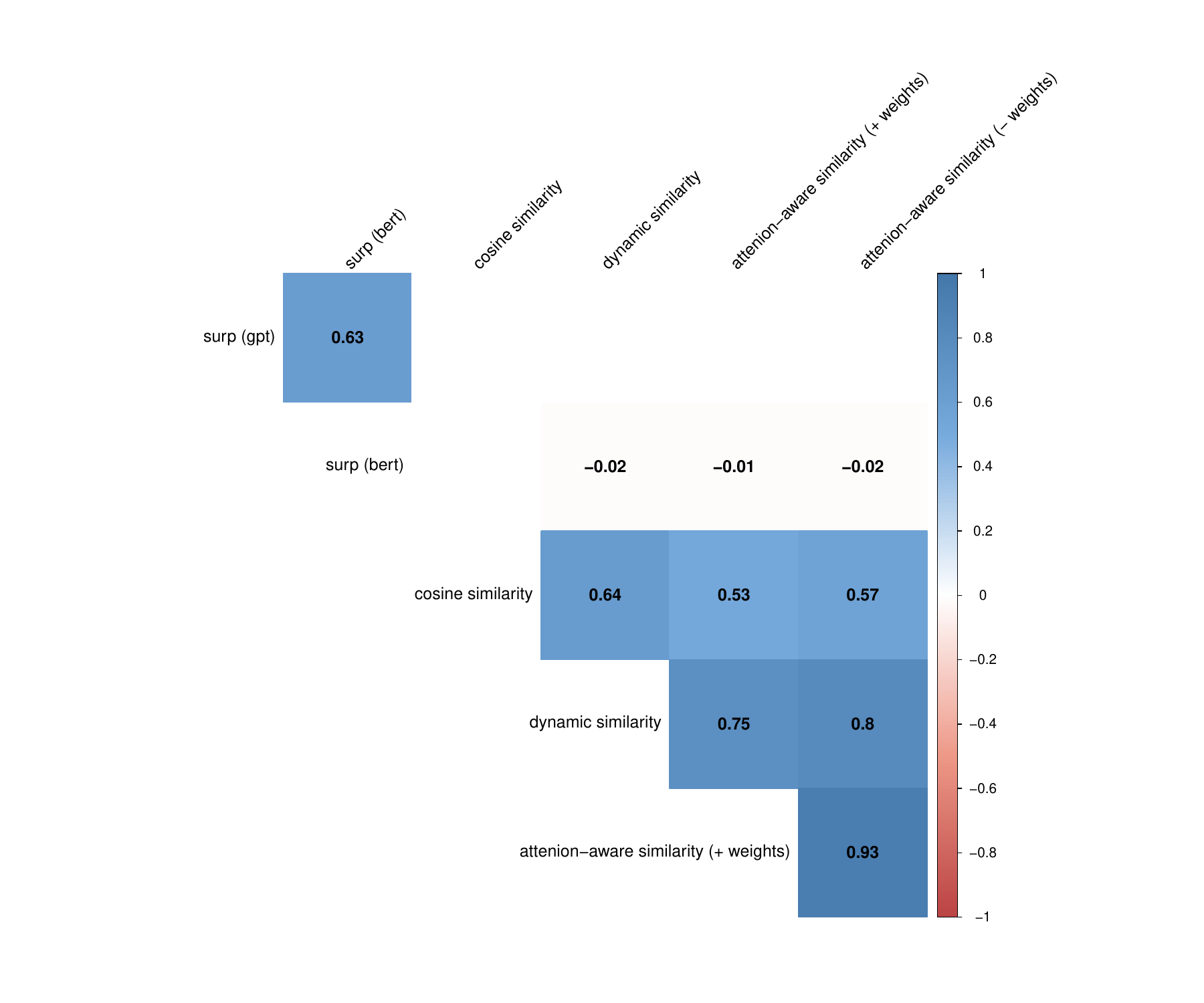}
	\caption{The Pearson correlations among various metrics of our interest}
	\label{fig:corr}
\end{figure}

Fig. \ref{fig:corr} presents a correlation matrix, elucidating the relationships between multiple metrics, including ``surp (surprsial computed by gpt)'', ``surp (surprisal computed by BERT)'', ``cosine similarity'', ``dynamic similarity", and ``attention-aware semantic relevance'', both with and without weights. This matrix is instrumental in revealing the degree of correlation between these metrics, providing insight into how similarly they correlate with one another across different dimensions or datasets. The matrix highlights notable positive correlations, especially for ``attention-aware similarity (+ weights)'', and underscores significant correlations between ``cosine similarity'' and ``dynamic similarity'' (attention-aware similarity without expectations). 
It indicates that surprisal metrics are distinct from semantic-based metrics we proposed in the present study. 

	\section{The predictability of word stroke count}

A number of studies have focused on examining how the stroke count of character influenced the processing of characters or word recognition in Chinese (\citealp{chen1996functional}; \citealp{ma2015character}; \citealp{liversedge2014effect}; \citealp{liu2002use}). However, the effect of word stroke count on word processing has been largely ignored.  
We are interested in exploring whether \texttt{stroke count of word} may help better predict the word processing in reading Chinese naturalistic discourse, compared to \texttt{stroke count of character} and \texttt{word length}.

To this end, we fitted three groups of GAMM models (nine models) to analyze the effect of stroke count of word as predictors of three dependent variables  \textit{first duration}, \textit{gaze duration} and \textit{total duration} from the same eye-movement dataset on Chinese reading \citep{zhang2022database}. The first group uses the GAMM fittings including \textit{word stroke count} and \textit{log(word frequency)}, as well as \textit{experiment} as a random effect. The GAMM fittings are formalized as: \textit{log(duration) $\sim$ s(log(word frequency)) + s(stroke count of word, experiment, bs=``fs'', m = 1)} (Here ``s''= smooth, \texttt{fs} = random smooths adjust the trend of a numeric predictor in a nonlinear way, and it cover the function of random intercept and random slope). In order to test the predictability of the stroke count of the first character in a word, we used the similar GAMM fittinsg in the second group: \textit{log(duration) $\sim$ s(log(word frequency)) + s(stroke count of 1st character, experiment, bs= ``fs'', m=1)}. \textit{`Stroke count of the first character'} is the stroke count of the first character in a word, and is similar to `stroke count of an individual character' in the relevant studies. Because \textit{word length} is nearly a binary factor,\textit{word length} cannot be processed with the smooth function, and the random variable cannot be applied in it with random smooth. The third group was formulated as:\textit{log(duration) $\sim$ s(log(word frequency)) +  word length + s(experiment, bs=``re'')}. 
For each duration, we conducted three types of GAMM fitting, resulting in a total of nine GAMM fittings carried out in our implementation.

We analyzed the \textit{p}-values of the independent variable and considered a significance level of $alpha$ \textit{p} < 0.01 to determine statistical significance. The results show that the variables of stroke count were significant in all GAMM fittings. The \textit{stroke count of the first character in word} also predicted all types of durations. The significant effects can be also seen in word length. 
Comparisons are established using consistent data point numbers (n = 76727) and identical elements for each GAMM fitting.  The calculation of $\Delta$\texttt{AIC} values involves subtracting the \texttt{AIC} of the base GAMM model (without the information on the metric of our interest) from the \texttt{AIC} of a full GAMM model. Smaller $\Delta$\texttt{AIC} values indicate a better fit for the GAMM model, suggesting superior performance for the metric used. The results are shown in Table \ref{tab:compare0}. In Table \ref{tab:compare0}, it is evident that the stroke count of a word outperforms either the stroke count of the first character or word length in predicting any of three types of duration. In general, the stroke count of word exhibited better overall performance compared to the other two metrics. 
In this way, \texttt{stroke count of word} could work as a control predictor in investigating Chinese reading compared with \texttt{word length} from the perspectives of informativeness and statistical assessment. That is why we employed \texttt{stroke count of word} as a control predictor in the present study to investigate the effects of the computational metrics we proposed. The following discusses why `` stroke count of word'' could act as a control predictor in Chinese reading.

\begin{table}
	\centering
	\caption{$\Delta$\texttt{AIC} for GAMM fittings with stroke count (n = 76727)}
	\scalebox{0.71}{
		\begin{tabular}{l *{3}{c}}
			\toprule
			\textbf{GAMM fittings} & {\textbf{log(First Duration)}} & {\textbf{log(Gaze Duration)}} & {\textbf{log(Total Duration)}} \\
			\midrule
			word length (character number) & -3388 & -5495.3 & -8976.49 \\
			stroke count of the 1st character & -6344 & -4231.4 & -2981.82 \\
			\textcolor{cyan}{stroke count of word} & \textcolor{cyan}{-3306.5} & \textcolor{cyan}{-12291.4} & \textcolor{cyan}{-11867.76} \\
			\bottomrule
		\end{tabular}
	}
	\label{tab:compare0}
\end{table}

\section{Control predictor of word stroke count} 
In a study by \citet{li2014reading}, the mixed effect of stroke count for individual characters in sentence reading was examined. \citet{li2020integrated} demonstrated that word length had a significant influence on word processing during Chinese naturalistic discourse reading. Building on these findings we have introduced in the relevant studies on word length and stroke count of character in Chinese reading, the current study finds that words with more strokes are associated with longer fixation durations, and this effect is further modified by word frequency.

The results show that readers of Chinese tend to fixate longer on words with more strokes, suggesting that the number of strokes affects the processing difficulty of word. One reason for this effect is that words with more strokes require more visual processing time to identify and process. This is because there are more visual features to process and integrate, such as the shape and orientation of each stroke, as well as the relationship between strokes within a word. As a result, readers tend to take more time to process the word fully. Moreover, \textit{word length} (i.e., the number of characters in a Chinese word) could be a reliable predictor of eye movements during reading in Chinese. However, the eye-tracking database we used provides limited information on word length, with ``1'' and ``2'' comprising almost half of the words respectively (i.e., 51\% vs. 46\%; ``3''--1.5\%, ``4''--0.9\%), making it difficult to use word length given the lack of variation. The reason for this is that word length for Chinese words is actually a binary factor variable. In this case, due to its binary feature, ``word length'' could play a role of random variable in regression models. On the other hand, \textit{word stroke count} ranges from 1 to 50, with the majority of data evenly distributed between 3 and 22. Therefore, \textit{word stroke count} may be a more informative predictor of eye movements in Chinese from the statistical perpsective, allowing it become a control predictor. However, the information of word length as group factor will greatly facilitate the predictability of word stroke of word.  

We further discuss about this. In written Chinese, characters are the basic visual units, and some characters function as standalone words. Researchers often control the number of character strokes in studies related to Chinese reading. However, exploring the stroke count of words can also provide valuable insights into the processing of Chinese words. It is evident that the stroke count of a word plays a significant role in influencing word processing during Chinese naturalistic discourse reading for three reasons. First, words with more strokes tend to require more time and cognitive effort to read and process compared to words with fewer strokes. The stroke count of word serves as an indicator of the overall visual complexity of the word as a whole. For instance, words with higher stroke counts contain more visually complex information, which impacts the speed and accuracy of word recognition and comprehension by readers. Second, in Chinese, words are generally composed of one or two characters, with two-character words exhibiting varying degrees of transparency between the meanings of the individual characters and the overall compound. Particularly for lower frequency words that may not be immediately recognized as a whole, the visual complexity of characters in terms of strokes appears to be a prominent factor analogous to the effects of character length observed in languages such as English. Third, in the eye-tracking corpus used in the current study, nearly half (49\%) of the words are composed of more than one character, making the total stroke count crucial in predicting how these multi-character words are processed. The character stroke count based on one-character words does not seem to adequately represent half of the words in the corpus. Indeed, the stroke count of a word can effectively represent all types of words. Through the comparison of GAMM fittings, we can confidently confirm the advantageous role of word stroke count in word processing.

Additionally, `word length' may play a role in predicting word processing during Chinese naturalistic discourse reading. For instance, word length has been shown to be a reliable predictor of certain eye-movement measures, such as gaze durations \citep{xiong2022multitask}. Despite of the importance of word length, \citet{fan2022eye} found that the effect of frequency did not significantly affect the probability of landing on a particular character in Chinese reading, while the number of strokes in the character did have a significant influence. We assume that the information provided by word length in Chinese is relatively limited from the statistic perspective. In English, `word length' typically falls within the range of 1 to 16 alphabets and is often used alongside word frequency as a control predictor in reading and eye-tracking studies. However, in Chinese, `word length' is usually restricted to a range of 1 to 4 characters. The concept of word length in Chinese is akin to the component number in compound words in English. Hence, we propose that the ``stroke count of a word'' could serve as an effective predictor of word processing in Chinese, complementing the predictive power of word length.  

In short, word length and stroke count of word are important factors that influence different aspects of the reading process in Chinese. Word length aids in processing character density, peripheral vision, and word segmentation and facilitates quicker lexical access, contributing to reading speed. On the other hand, the stroke count of word reflects overall visual complexity, impacting fixation durations and cognitive engagement during reading. 
Despite this, ``stroke count of word'' could better act as a control predictor in Chinese naturalistic reading in comparison with ``word length''. 

\end{document}